\def\year{2022}\relax
\documentclass[letterpaper]{article} 
\usepackage{aaai22}  
\usepackage{times}  
\usepackage{helvet}  
\usepackage{courier}  
\usepackage[hyphens]{url}  
\usepackage{graphicx} 
\usepackage{natbib}  
\usepackage{caption} 
\DeclareCaptionStyle{ruled}{labelfont=normalfont,labelsep=colon,strut=off} 
\usepackage{algorithm}
\usepackage{algorithmic}

\usepackage{newfloat}
\usepackage{listings}

\usepackage{amsmath}
\usepackage{amsfonts}
\usepackage{multirow}

\floatstyle{ruled}
\newfloat{listing}{tb}{lst}{}
\floatname{listing}{Listing}
\title{Self-Supervised Video Representation Learning by Video Incoherence Detection}
\author{
    Haozhi Cao,\equalcontrib\textsuperscript{\rm 1}
    Yuecong Xu,\equalcontrib\textsuperscript{\rm 2}
    Jianfei Yang,\textsuperscript{\rm 1}\\
    Kezhi Mao,\textsuperscript{\rm 1}
    Lihua Xie, \textsuperscript{\rm 1}
    Jianxiong Yin, \textsuperscript{\rm 3}
    Simon See \textsuperscript{\rm 3}
}
\affiliations{
    \textsuperscript{\rm 1}School of Electrical and Electronic Engineering, Nanyang Technological University, Singapore,\\
    \textsuperscript{\rm 2}Institute for Infocomm Research, A*STAR, Singapore,
    \textsuperscript{\rm 3}NVIDIA AI Tech Centre,\\
    \{haozhi001, xuyu0014, yang0478\}@e.ntu.edu.sg,
    \{ekzmao, elhxie\}@ntu.edu.sg,
    \{jianxiongy, ssee\}@nvidia.com,
}

\usepackage{bibentry}

\begin{document}
\hbadness=1000000000
\vbadness=1000000000
\hfuzz=10pt

\setlength{\abovedisplayskip}{2pt}
\setlength{\belowdisplayskip}{2pt}
\setlength{\textfloatsep}{10pt plus 1.0pt minus 2.0pt}
\setlength{\floatsep}{6pt plus 1.0pt minus 1.0pt}
\setlength{\intextsep}{6pt plus 1.0pt minus 1.0pt}
\setlength{\abovedisplayshortskip}{0pt}
\setlength{\belowdisplayshortskip}{0pt}

\maketitle

\begin{abstract}
   This paper introduces a novel self-supervised method that leverages incoherence detection for video representation learning. It roots from the observation that visual systems of human beings can easily identify video incoherence based on their comprehensive understanding of videos. Specifically, the training sample, denoted as the incoherent clip, is constructed by multiple sub-clips hierarchically sampled from the same raw video with various lengths of incoherence between each other. The network is trained to learn high-level representation by predicting the location and length of incoherence given the incoherent clip as input. Additionally, intra-video contrastive learning is introduced to maximize the mutual information between incoherent clips from the same raw video. We evaluate our proposed method through extensive experiments on action recognition and video retrieval utilizing various backbone networks. Experiments show that our proposed method achieves state-of-the-art performance across different backbone networks and different datasets compared with previous coherence-based methods.
\end{abstract}

\section{Introduction}
Fully supervised learning has achieved great success in video representation learning during the past decade. However, its outstanding performance significantly relies on a large amount of labeled data, whose annotation is resource-expensive and time-consuming. Additionally, fully supervised methods are designed to extract task-specific representation, suffering from poor transfer-ability and generalization. To address these issues, recent works have paid more attention to self-supervised learning, which aims to extract generalized representation from more accessible unlabeled data on the Internet.

\begin{figure}[t]
    \centering{
    \includegraphics[width=.73\linewidth]{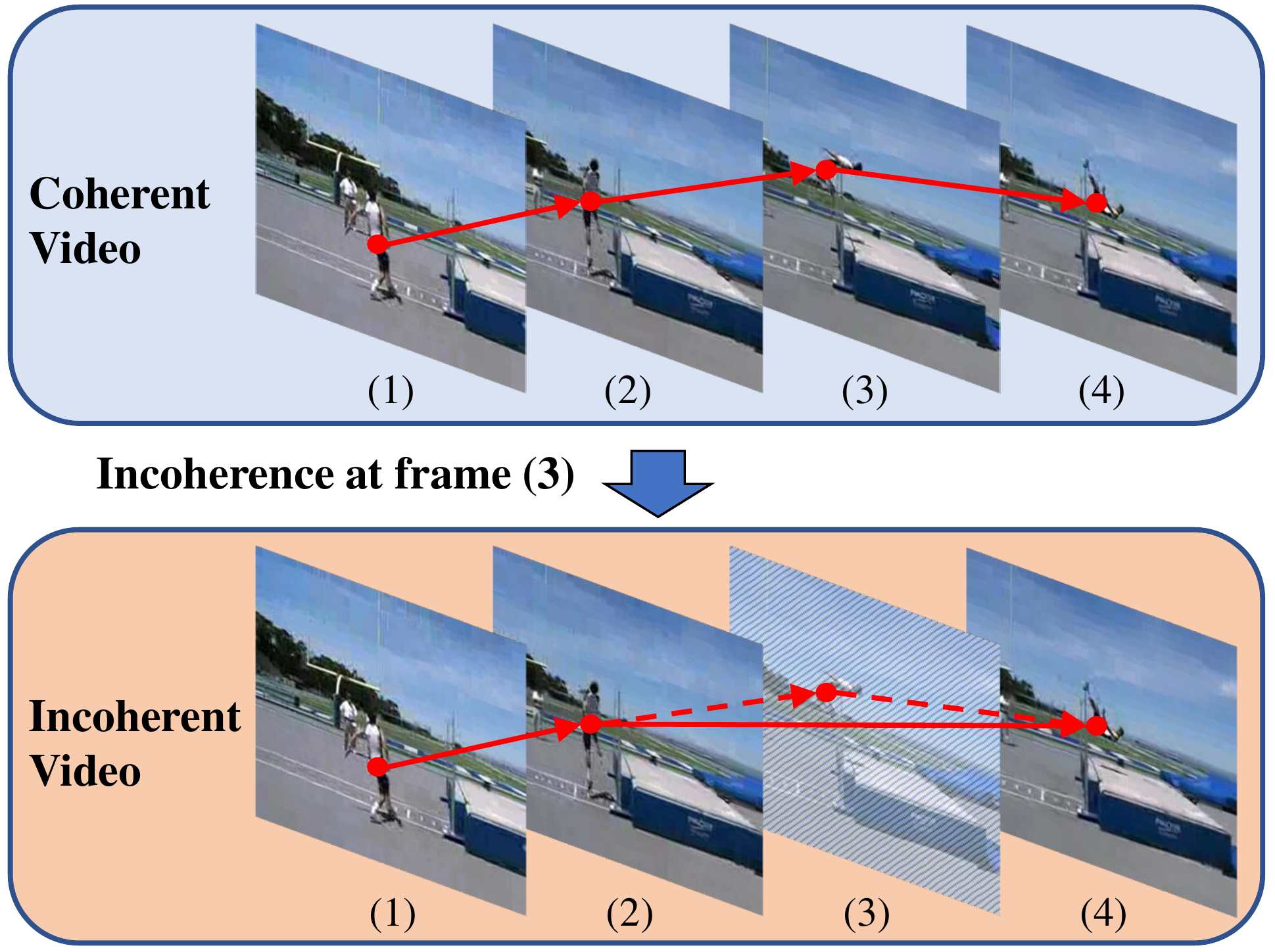}}
    \caption{Illustration about how video incoherence affects motion information. Both videos demonstrate the action ``High Jump". The incoherence caused by losing frame (3) leads to a distortion of athlete motion between frame (2) and frame (4) in the incoherent video, which is incompatible with our understanding of ``High Jump". This observation suggests that incoherence detection requires a comprehensive understanding of videos.}
    \label{Figure:Intro-1}
\end{figure}

Typically, the core of self-supervised methods is to design a pretext task where the network is driven to learn representation through characteristics of unlabeled data. Existing self-supervised methods can be categorized into two main types: (i) dense prediction and (ii) spatio-temporal reasoning. 
Methods using dense prediction require the network to predict parts of low-level representation, such as future frames \cite{Carl2016Generating,Nitish2015frameprediction} and optical flows \cite{Gan2018Geometry}. While outstanding performance can be achieved, they usually require hand-crafted features (e.g. optical flow \cite{Gan2018Geometry}) or complicated computation process \cite{Yuan2020CGMUL,Han2020MemDPC}, leading to an expensive cost of time and resources. To improve the efficiency, recent spatio-temporal reasoning methods, such as clip order prediction \cite{xu2019VCOP,fernando2017O3D,misra2016shuffle,lee2017sortsequence}, video speed prediction \cite{wang2020PP,jenni2020RTT,Yao_2020PRP} and spatio-temporal statistic prediction \cite{wang2019PMAS}, tend to learn the high-level spatio-temporal correlations of raw videos. Methods based on clip order prediction attempt to leverage the video coherence for representation learning, where supervision signal is generated from frame order disruption. In this paper, we propose an incoherence detection method to leverage video coherence for video representation learning from a new perspective. 

Intuitively, our visual systems can easily identify the incoherence of videos (e.g. loss frames caused by connection latency), since we can speculate the abnormal motion based on our understanding of videos. In this case, the incoherence can be viewed as the noise to motion information. To detect incoherence, the network requires a comprehensive understanding of videos that motivates our paper. For example, as illustrated in Figure~\ref{Figure:Intro-1}, we can easily justify whether there exists some incoherence between frame (2) and frame (4). This is because given previous frames (1-2), we can deduce that the athlete should be leaping over the bar in the next frame. Yet given frames (1-2) where the athlete remains on the left side of the bar, the athlete in the frame (4) suddenly appears on the right side without the process of leaping, which is incompatible with our deduction. This bi-directional reasoning of video contents could be an effective supervision signal for the network to learn high-level representation of videos. 

Inspired by this observation, we propose a simple-yet-effective method called Video Incoherence Detection (VID) for video representation learning in a self-supervised manner. Each training sample is generated as an incoherent clip constructed by multiple sub-clips from the same raw video. Specifically, sub-clips are hierarchically sampled from the raw video given the random incoherence location and length. The incoherent clip is then constructed as the concatenation of sub-clips along the temporal dimension. Different from previous coherence-based methods \cite{xu2019VCOP,fernando2017O3D,misra2016shuffle,lee2017sortsequence} which undermine temporal orders, VID preserves the sequential relationship of the raw video during the generation. The network can therefore learn temporal representation for incoherence detection.

Given the incoherent clips as input, the network is trained to detect the incoherence by two novel pretext tasks that predict the location and length of incoherence, denoted as Incoherence Location Detection (LoD) and Incoherence Length Detection (LeD), respectively. Moreover, we introduce the Intra-Video Contrastive Learning (ICL) as an additional optimization objective to maximize the mutual information between different incoherent clips from the same raw video.

In summary, our contributions are three-fold. Firstly, motivated by the fact that detecting incoherence requires semantic understanding, we propose a simple-yet-effective self-supervised method, called Video Incoherence Detection (VID), utilizing a single temporal transformation method for video representation learning. Secondly, Incoherence Location Detection (LoD) and Incoherence Length Detection (LeD) are proposed to learn spatio-temporal representation by detecting incoherence while avoiding shortcuts. Thirdly, we introduce Intra-Video Contrastive Learning (ICL) to maximize the mutual information between incoherent clips from the same video. Extensive experiments show that our VID achieves state-of-the-art performance on action recognition and video retrieval compared with previous coherence-based methods.


\section{Literature review}

\textbf{Self-supervised learning.}
To leverage more accessible unlabeled data on the Internet, recent methods pay more attention to self-supervised learning. Self-supervised learning stems from the previous works \cite{Caruana1997Promoting,ando2005framework} and has been widely explored in images \cite{wu2018unsupervised,Hjelm2019MIE,Misra_2020_PIR} or natural language \cite{devlin2018bert,lan2019albert}. Early works have expanded self-supervised methods from other domains to videos, e.g. DPC \cite{han2019DPC} inspired by CPC \cite{oord2018representation} in image domain and \cite{sun2019videobert,sun2019learning} inspired by BERT \cite{devlin2018bert}. 

Recent self-supervised methods for video representation learning can be categorized into two types, including dense prediction and spatio-temporal reasoning. Methods based on dense prediction  \cite{Carl2016Generating,Gan2018Geometry,han2019DPC,Han2020MemDPC,Nitish2015frameprediction,Yuan2020CGMUL} require network to predict the low-level information of videos. \cite{Carl2016Generating,Nitish2015frameprediction} proposed to learn video representation by predicting future frames whose foreground and background are generated from independent streams. To leverage video information of multi-modalities, some previous works proposed to generate supervision signal through the input of multi-modalities, such as 3D videos \cite{Gan2018Geometry} and RGB-D data \cite{luo2017unsupervised}. 

Instead of directly predicting low-level information, methods based on spatio-temporal reasoning are proposed to generate supervision signals as correlations or characteristics of videos. compared with dense prediction, previous spatio-temporal reasoning methods require dedicate pretext tasks, such as temporal order prediction \cite{fernando2017O3D,xu2019VCOP,lee2017sortsequence,misra2016shuffle,Kim2019STpuzzle} and video speed prediction \cite{wang2020PP,Yao_2020PRP,jenni2020RTT}. Inspired by the sequential relationships of videos, previous works \cite{fernando2017O3D,lee2017sortsequence,misra2016shuffle} attempted to predict or identify the correct frame order given clips shuffled along the temporal dimension. \cite{xu2019VCOP} further applied the order prediction method with 3D-CNN and \cite{Kim2019STpuzzle} expanded the order prediction to the spatial dimension. On the other hand, recent methods \cite{Yao_2020PRP,wang2020PP,jenni2020RTT} proposed to extract effective representation by predicting the speed of videos. Specifically, \cite{Yao_2020PRP,wang2020PP} combined the speed prediction task with re-generation and contrastive learning, respectively. \cite{jenni2020RTT} achieved state-of-the-art performance by recognizing various temporal transformations under different speed. Inspired by humans' sensitivity towards incoherence in videos, we argue that video incoherence detection requires semantic understanding of video contents, which can be explored to learn effective video representation.

\textbf{Contrastive learning.}
Contrastive learning has been proven to be an effective optimization objective in self-supervised learning. For image representation learning, multiple methods \cite{Misra_2020_PIR,Hjelm2019MIE,Phillip2019MM,wu2018unsupervised} proposed to learning effective image representation by using contrastive learning. Inspired by the success of contrastive learning in images, recent methods \cite{Dwibedi2019TCC,wang2020PP,he2020momentum,lorre2020temporal,yao2021seco} have been proposed to leverage contrastive learning for video representation learning. The basic idea is to maximize the mutual information between positive pairs. For instance, \cite{Dwibedi2019TCC,wang2020PP} attempted to align spatio-temporal representation of the same action or same context. Recently, \cite{yao2021seco} conducted contrastive learning from spatial, spatio-temporal and sequential perspectives. In this work, we utilize intra-video contrastive learning to maximize the mutual information between different incoherent clips from the same video.

\section{Proposed methods}
Coherence is one of the crucial properties of videos. Natural videos are formed by sets of frames coherently observed. Our visual systems can easily identify incoherence caused by loss frames within a video clip, which demonstrates that the detection of incoherence would require a semantic understanding of videos. This motivates us to design a self-supervised method by leveraging incoherence detection for video representation learning.

In this work, we propose to extract effective spatio-temporal representation by Video Incoherence Detection (VID) based on a single temporal transformation in a self-supervised manner. We first illustrate how to generate incoherent clips from raw videos. Based on these generated clips, Location Detection (LoD), Length Detection (LeD) and Intra-Video Contrastive Learning (ICL) are proposed for self-supervised learning in details. To clarify the whole learning procedure, we summarize the overall learning objective and framework of VID in Section~\ref{Section:Method-Structure}.

\subsection{Generation of incoherent video clips}\label{Section:Generation}
\begin{figure}[t]
    \centering{
    \includegraphics[width=1.\linewidth]{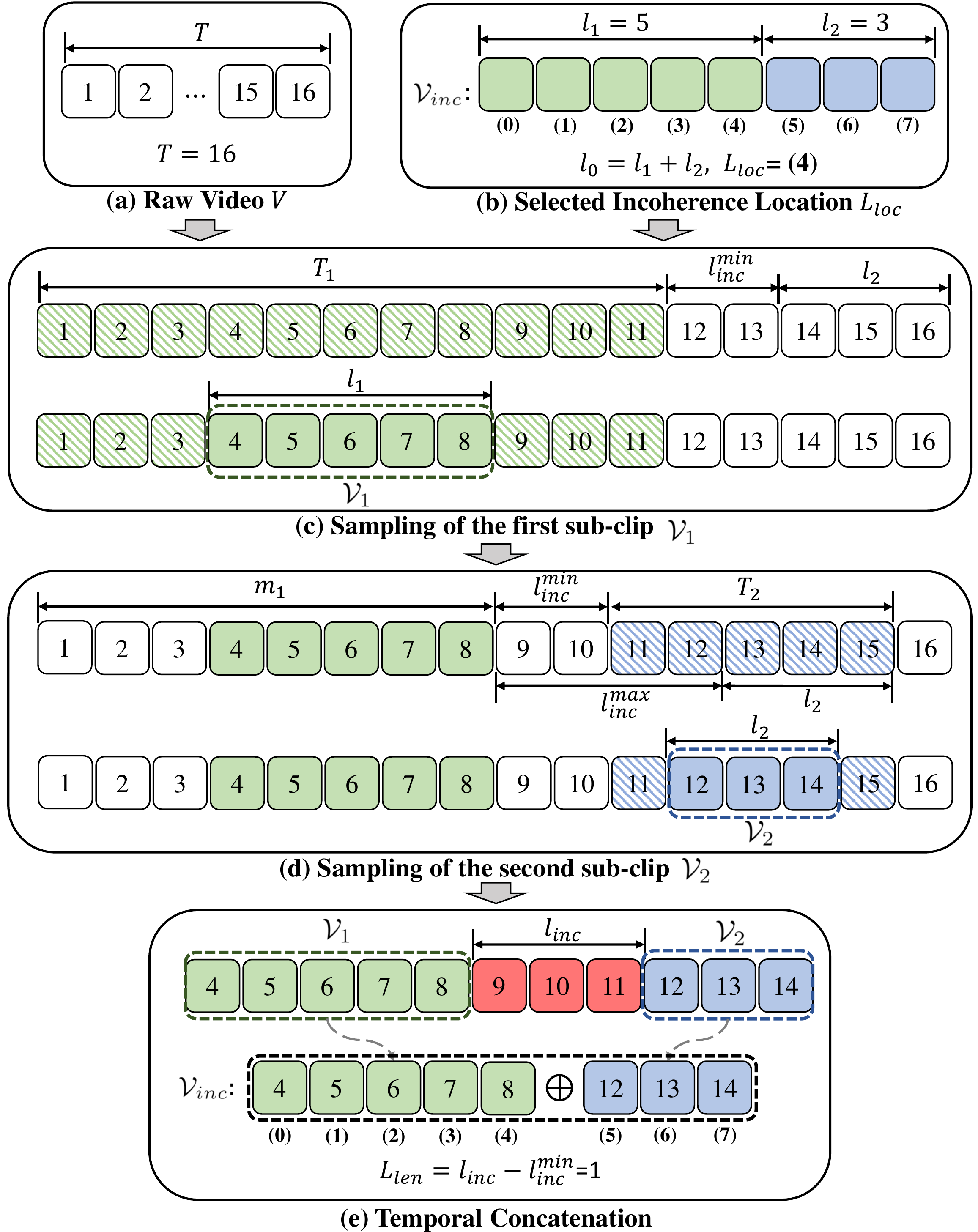}}
    \hfill
    \caption{Generation process of the incoherent clip $\mathcal{V}_{inc}$. Indices 1-16 in square denote the frame indices in the raw video $V$ while indices (0-7) denote the relative frame indices in $\mathcal{V}_{inc}$. Squares in shadow and color refer to the sample range and sampled frames for the correspondent sub-clip, respectively. $\mathcal{V}_{inc}$ is generated as the concatenation of $\mathcal{V}_1$ and $\mathcal{V}_2$ along the temporal dimension as shown in (e).}
    \label{Figure:Method-1}
\end{figure}

To utilize VID, we first generate incoherent clips from raw videos. Given a raw video $V$, the incoherent clip $\mathcal{V}_{inc}$ is constructed by $k$ sub-clips $\mathcal{V}_1,\mathcal{V}_2,...,\mathcal{V}_k$ sampled from $V$ with a certain length of incoherence between each other. The location $L_{loc}$ and length $l_{inc}$ of incoherence are both randomly generated. The length of incoherence $l_{inc}$ between sub-clips is limited within the range of:
\begin{equation}\label{Equation:LengthConstraint}
    l_{inc} \in [l_{inc}^{min}, l_{inc}^{min}+1,...,l_{inc}^{max}],
\end{equation}
 where $l_{inc}^{min}, l_{inc}^{max}$ are both hyper-parameters indicating the upper and lower bounds of the incoherence length, respectively. The purposes of this constraint are two-fold. Firstly, the constraint on $l_{inc}$ is necessary for our Incoherence Length Detection (LeD). Secondly, the constraint on incoherence length prevents the incoherence between sub-clips from being either too vague or too obvious, which avoids learning trivial solutions. For simplicity, we take the case where $\mathcal{V}_{inc}$ is constructed by two sub-clips as an example to thoroughly illustrate the generation process of incoherent clips as shown in Figure~\ref{Figure:Method-1}.

\textbf{Selection of incoherence location.}
The incoherence location $L_{loc}$ refers to the relative concatenation location between two sub-clips. Formally, given the desired length of incoherent clips $l_0$, the location of incoherence $L_{loc}$ is uniformly selected as:
\begin{align}
    & l_1 \in \{ 1,2,...,l_0-1 \}, \ \ l_2 = l_0 - l_1, \\
    & L_{loc} = l_1 - 1,
\end{align}
where $l_1$, $l_2$ are the length of $\mathcal{V}_1$, $\mathcal{V}_2$ as illustrated in Figure~\ref{Figure:Method-1}(b), where squares in different colors refer to allocated frame positions for different sub-clips in $\mathcal{V}_{inc}$. $L_{loc}$ is the relative location of incoherence where sub-clips concatenate as well as the label for the following LoD task.

\begin{figure*}[!ht]
    \centering{
    \includegraphics[width=0.9\textwidth]{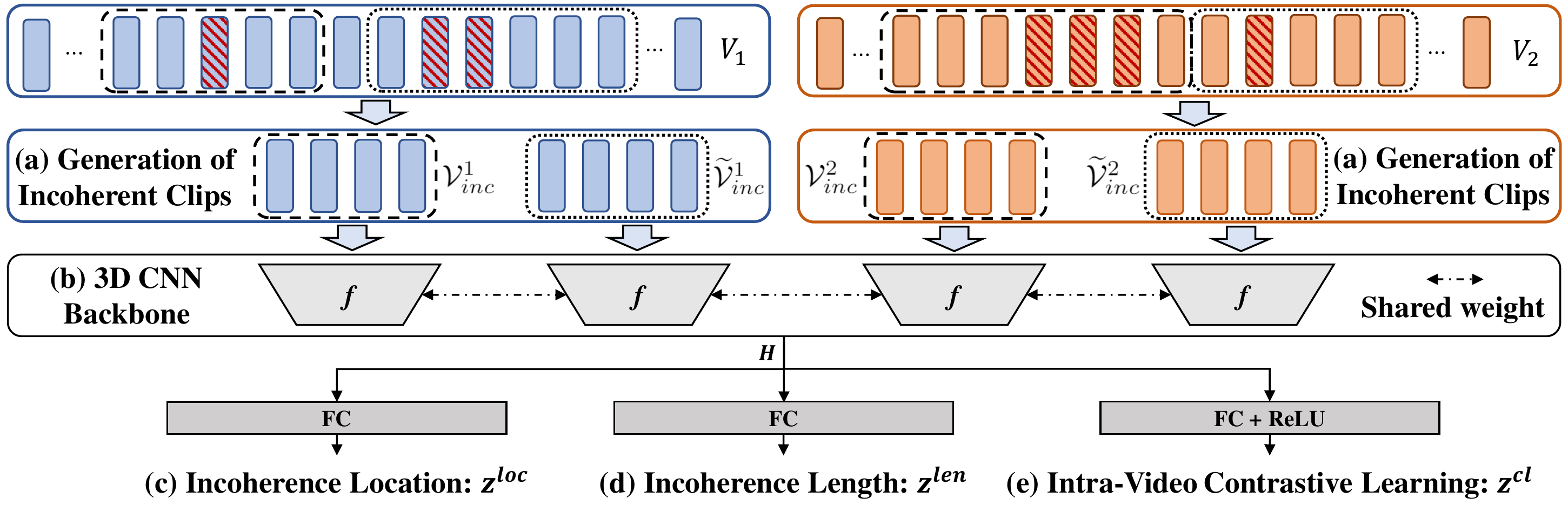}}
    \hfill
    \caption{The structure of our proposed VID method. The first row indicates two raw videos $V_1$, $V_2$. Two incoherent clips are generated from each raw video. Subsequently, they are fed into a single 3D CNN backbone. The extracted high-level representation $H$ then passes to three different linear or non-linear layers to perform three different sub-tasks, including Incoherence Location Detection (LoD), Incoherence Length Detection (LeD) and Intra-Video Contrastive Learning (ICL).}
    \label{Figure:Method-2}
\end{figure*}

\textbf{Hierarchical selection of sub-clips.}
Given the sub-clip lengths $l_1,l_2$, sub-clips $\mathcal{V}_1$, $\mathcal{V}_2$ are hierarchically sampled from the raw video $V$ with incoherence between each other. The incoherent clip $\mathcal{V}_{inc}$ is generated as the temporal concatenation of $\mathcal{V}_1$, $\mathcal{V}_2$.
While previous work \cite{wang2020PP} proposed to sample frames by looping over the raw video, such strategy is not compatible with our proposed VID since it could introduce unexpected incoherence when looping from the end to the start of the video. Instead, to preserve the sequential relationship of the raw video, we propose a hierarchical sampling strategy that maximizes the sample range of each sub-clip while satisfying Equation~\ref{Equation:LengthConstraint}.

Illustrated as the upper row in Figure~\ref{Figure:Method-1}(c), given the raw video $V$ of the length $T$, the sample range $T_1$ of the first sub-clip $\mathcal{V}_1$ is determined by reserving sufficient frames for subsequent sub-clips. In this way, the sub-clip $\mathcal{V}_2$ can be sampled from the rest of the raw frames wherever the $\mathcal{V}_1$ locates in $T_1$, which preserves the sequential relationship and satisfies the constraint in Equation~\ref{Equation:LengthConstraint}. Formally, given $l_2$ and $l_{inc}^{min}$, the sample range $T_1$ is computed as:
\begin{align}
    t_1^{min} &= 1, \\
    t_1^{max} &= T-l_{inc}^{min}-l_2, \\
    T_1 &= \{ t_1^{min}, t_1^{min}+1,...,t_1^{max} \},
\end{align}
where $t_1^{min}$, $t_1^{max}$ are the lower and upper bound of the range $T_1$. Given the range $T_1$, $\mathcal{V}_1$ is uniformly sampled as $\mathcal{V}_1\in T_1$ illustrated as the lower row in Figure~\ref{Figure:Method-1}(c). 

Hierarchically, the range of the second sub-clip $T_2$ is decided by the sampled sub-clip $\mathcal{V}_1$ and the range of $l_{inc}$ in Equation~\ref{Equation:LengthConstraint}. As shown in the upper row of Figure~\ref{Figure:Method-1}(d), given the raw frame index of the last frame in $\mathcal{V}_1$ denoted as $m_1=\max(\mathcal{V}_1)$, the sample range $T_2$ is computed as:
\begin{align}
    t_2^{min} &= m_1+l_{inc}^{min}+1, \\
    t_2^{max} &= \min (m_1+l_{inc}^{max}+l_2, T), \\
    T_2 &= \{ t_2^{min}, t_2^{min}+1,...,t_2^{max} \},
\end{align}
where $t_2^{min}$, $t_2^{max}$ are the upper and lower bounds which ensure that $l_{inc}$, the length of incoherence between $\mathcal{V}_2$ and $\mathcal{V}_1$, always satisfies the constraints in Equation~\ref{Equation:LengthConstraint}. Similar to $\mathcal{V}_1$, the second clip $\mathcal{V}_2$ is uniformly sampled as $\mathcal{V}_2\in T_2$, as in the lower row of Figure~\ref{Figure:Method-1}(d).

Given the sub-clips $\mathcal{V}_1,\mathcal{V}_2$, the incoherent clip $\mathcal{V}_{inc}$ and its label $L_{len}$ for Incoherence Length Detection (LeD) task are generated as:
\begin{align}
    \mathcal{V}_{inc} &= \mathcal{V}_1 \oplus \mathcal{V}_2, \\
    l_{inc} &= \min(\mathcal{V}_2)-\max(\mathcal{V}_1), \\
    L_{len} &= l_{inc}-l_{inc}^{min},\label{Equation:LableLen}
\end{align}
where $\oplus$ indicates the concatenation of two sub-clips $\mathcal{V}_1$, $\mathcal{V}_2$ along the temporal dimension.

\subsection{Optimization objectives}\label{Section:ID}
We propose two novel self-supervised tasks, including Incoherence Location Detection (LoD) and Incoherence Length Detection (LeD) to detect the incoherence in incoherence clips while maximizing the mutual information between different incoherent clips from the same raw video. Specifically, given an incoherent clip $\mathcal{V}_{inc}$, the high-level representation is first extracted as $h=f(\mathcal{V}_{inc})$ where $f(\cdot)$ denotes the encoder. Given the representation $h$, the optimization objectives of VID include three components: Incoherence Location Detection (LoD), Incoherence Length Detection (LeD) and Intra-Video Contrastive Learning (ICL).

\textbf{Incoherence Location Detection (LoD).}
Given the high-level representation $h$ and its location label $L_{loc}$, the network is required to predict the location of incoherence in $\mathcal{V}_{inc}$. This is mainly inspired by humans' sensitivity towards the loss frames within video clips. The network is driven to identify the abnormal motion caused by incoherence, which encourages the network to learn semantic representation about videos. The LoD task is formulated as a single-label classification problem. Given the representation $h$ and label $L_{loc}$, the network is optimized by the cross-entropy loss illustrated as:

\begin{equation}\label{Equation:LoC}
    l_{LoD} = -\sum_{i=0}^{l_0-1}y^{loc}_i \log(\frac{\exp{(z^{loc}_i)}}{\sum_{j=0}^{l_0-1}\exp{(z^{loc}_j})}),
\end{equation}

where $z^{loc}\in\mathbb{R}^{l_0-1}$ is the output of fully-connected layers $\phi^{loc}(\cdot)$ given the representation $h$ as input. $y^{loc}\in\mathbb{R}^{l_0-1} $ is the ground-truth label vector whose element at $L_{loc}$ equals 1 while the rest to 0. In practice, given a mini-batch of representation $Z^{loc}$, Equation~\ref{Equation:LoC} is applied to each representation in $Z^{loc} \in \mathbb{R}^{N\times (l_0-1)}$, where $N$ denotes the batch size. The $\mathcal{L}_{loc}$ loss is then calculated as the average of all losses of representation in $Z^{loc}$.

\textbf{Incoherence Length Detection (LeD).}
In addition to LoD, given the high-level representation $h$ and its corresponding label $L_{len}$ of Equation~\ref{Equation:LableLen}, the network is required to predict the length of incoherence. The proposed LeD task is designed as a regularization measure to avoid trivial learning. In some cases, it is possible that the incoherence locates at the period when the distribution of low-level representation intensively changes (e.g. intensive movement of the camera or sudden changes of light conditions). This could cause a distinct difference in low-level representation between sub-clips of the incoherent clip $\mathcal{V}_{inc}$, leading to trivial learning during LoD. compared with LoD which extracts semantic representation, our proposed LeD can be regarded as a simple yet challenging task that extracts additional temporal information by enforcing the network to deduce the length of incoherence with respect to the raw video. In practice, the accuracy of LeD is relatively low (with Top1 about 25\%), while our ablation experiment shows that it can bring noticeable improvement to the performance mainly because it can improve the robustness of VID towards intensive changes of low-level information.

Similar to LoD, the LeD task can also be formulated as a classification problem, where cross-entropy loss is utilized for optimization as:
\begin{equation}\label{Equation:LeD}
     l_{LeD} = -\sum_{i=0}^{\Delta l_{inc}}y^{len}_i \log(\frac{\exp{(z^{len}_i)}}{\sum_{j=0}^{\Delta l_{inc}}\exp{(z^{len}_j})}),
\end{equation}
where $z^{len}\in\mathbb{R}^{l_0-1}$ is the output of fully-connected layers $\phi^{len}(\cdot)$ given $h$ as input. $y^{len}\in\mathbb{R}^{\Delta l_{inc}}$ is the ground-truth label of incoherence length and $\Delta l_{inc}$ is the difference between the upper bound and lower bound of incoherence length. Similar to LoD, Equation~\ref{Equation:LeD} is also applied to each representation of the mini-batch $Z^{len}\in\mathbb{R}^{N\times\Delta l_{inc}}$, whose loss $\mathcal{L}_{LeD}$ is calculated as the average of all losses of $Z^{len}$.

\textbf{Intra-Video Contrastive Learning (ICL).}
Contrastive learning can effectively extract the mutual information between variously augmented samples from the same source. Recent works \cite{yao2021seco,wang2020PP} demonstrate its great potential, exceeding other self-supervised or even supervised methods. In this work, we include Intra-Video Contrastive Learning (ICL) as an extra optimization objective to maximize the mutual information between different incoherent clips from the same video. This is inspired by the fact that our visual systems extract mutual information from incoherent clips, and thus can still recognize the correct motions even under incoherent circumstances.

Formally, given a mini-batch of $N$ raw videos $\mathbf{V}=\{ V_1,V_2,...,V_N \}$, two incoherent clips are randomly generated as demonstrated in Section~\ref{Section:Generation} for each raw video $V_i\in \mathbf{V}$. The incoherent clips from the same raw video $V_i$ are considered as positive pairs denoted as $\{\mathcal{V}_{inc}^i,\widetilde{\mathcal{V}}_{inc}^i\}$. The incoherent clips from different raw videos are regarded as negative pairs denoted as $\{ \mathcal{V}_{inc}^i,\mathcal{V}_{inc}^k \},k \neq i$. Each incoherent clip $\mathcal{V}_{inc}^i$ is then fed to the network $f(\cdot)$, forming the high-level representation $h_i$. The representation $h_i$ is subsequently fed to a fully-connected layer $\phi^{cl}(\cdot)$ followed by a non-linear ReLU activation, generating features $z_i^{cl}$. Provided with features of positive pairs $\{z_i^{cl}, \widetilde{z_i}^{cl}\}$ and features of negative pairs $\{z_i^{cl}, z_k^{cl}\},k\neq i$, the contrastive loss is computed as:
\begin{equation}
    \mathcal{L}_{ICL} = -\frac{1}{2N} \sum_{i=1}^{2N} \log(\frac{\exp{(s(z^{cl}_i,\widetilde{z_i}^{cl})})}{\exp{(s(z^{cl}_i,\widetilde{z_i}^{cl})})+\mathcal{D}(i))}),
\end{equation}
\begin{equation}
    \mathcal{D}(i)=\sum\nolimits_{k \neq i}\exp{(s(z^{cl}_i,z_k^{cl}))},
\end{equation}
where $s(u,v)=u^{\top}v/\left\|u\right\|\left\|v\right\|$ denotes the similarity between feature $u$ and $v$. $\mathcal{D}(i)$ is the summation of exponential similarity between features of negative pairs.

\subsection{Network structure and training}\label{Section:Method-Structure}
The overall network structure is illustrated as Figure~\ref{Figure:Method-2}. Given the unlabeled raw video, the incoherent clips are first generated as described in Section~\ref{Section:Generation}, where each raw video randomly generates two incoherent clips as shown in Figure~\ref{Figure:Method-2}(a). Subsequently, the batch of incoherent clips is fed to the encoder $f(\cdot)$ implemented as the 3D CNN backbone. The ultimate optimization objective is formulated as:
\begin{equation}
    \mathcal{L}=\alpha\mathcal{L}_{LoD}+\beta\mathcal{L}_{LeD}+\lambda\mathcal{L}_{ICL},
\end{equation}
where $\alpha$, $\beta$, $\lambda$ are the coefficients of three loss terms from the sub-tasks, respectively. 

\section{Experiments}
In this section, we present thorough experiments to justify our proposed VID. We first illustrate our experiment settings and subsequently justify our VID design through detailed ablation studies. Finally, VID is evaluated on two downstream tasks including action recognition and video retrieval in comparison with state-of-the-art methods. Code and visualization are available at the supplementary material.

\subsection{Experiment settings}\label{Sec:Exp-Settings}
\textbf{Datasets.}
We evaluate our VID across three action recognition datasets, including UCF101 \cite{soomro2012ucf101}, HMDB51 \cite{kuehne2011hmdb} and Kinetics-400 \cite{kay2017kinetics}. UCF101 is a widely used video dataset for action recognition, which contains 13,320 videos with 101 action categories. HMDB51 is a relatively smaller yet challenging dataset for action recognition. It includes about 7,000 videos with 51 action classes. Both UCF101 and HMDB51 are divided into three training and testing splits. 
Kinetics-400, denoted as K-400, is a large dataset for action recognition. It contains about 304,000 videos with 400 action classes collected from the online video platform YouTube. Same as the setting of prior work \cite{wang2020PP,jenni2020RTT}, we utilize the training split of Kinetics-400 and the training split 1 of UCF101 for self-supervised pre-training. The training split 1 of UCF101 and HMDB51 are utilized during fine-tuning for action recognition.

\textbf{Backbone networks.}
As for the 3D CNN backbones, to fairly compare our proposed methods with others \cite{wang2020PP,xu2019VCOP}, we utilize three different 3D CNN networks in our experiments, including C3D \cite{Tran2015C3D}, R3D \cite{hara2018can} and R(2+1)D \cite{tran2018closer}. The mentioned backbones have been widely used to evaluate self-supervised methods in previous research. Specifically, C3D \cite{Tran2015C3D} is constructed by direct extending 2D kernels of 2D CNN to 3D ones. R3D \cite{hara2018can} introduces the residual connections from 2D CNNs to 3D CNNs. Following previous works \cite{Yao_2020PRP,jenni2020RTT,Kim2019STpuzzle}, we utilize R3D-18 which is the 18-layer variant of R3D. R(2+1)D \cite{tran2018closer} proposes to replace the traditional 3D kernel with the combination of a 2D kernel and a 1D kernel for spatial and temporal feature extraction, respectively. In this work, we mainly conduct our experiments with R(2+1)D thanks to its superior performance compared with others.

\textbf{Augmentation and parameters.}
Following the setting of prior work \cite{jenni2020RTT,wang2020PP}, each incoherent clip includes 16 frames. The sampling for each sub-clip is 1 and the range of incoherence length is set as $l_{inc} \in \{3,4,...,10\}$. While pre-training on UCF101, we follow the setting in \cite{wang2020PP,alwassel2019self} which increases the epoch size from 9k to 90k with color jittering along the temporal dimension. Frames are resized to $128\times171$ and then randomly cropped to $112\times112$. The whole input clip is then flipped horizontally with a probability of $50\%$. The network is trained with a batch-size of 30. The stochastic gradient descent \cite{bottou2010SGD} is utilized for optimization with the weight decay set to 0.005 and the momentum set to 0.9. The coefficients of sub-tasks $\alpha$, $\beta$, $\lambda$ are empirically set to 1, 0.1, 0.1. The learning rate is initialized as 0.001 and divided by 10 every 6 epochs with a total training epoch of 18. 

\begin{table}[t]
    \centering
    \resizebox{.65\linewidth}{!}{
    \begin{tabular}{c|c|c|c}
    \hline
    \hline
    Method & Jittering & Range of $l_{inc}$ & UCF101(\%) \\
    \hline
    Random & \checkmark & - & 56.7 \\
    \hline
    \multirow{9}{*}{\centering VID}
    & \checkmark & $[2,10]$ & 76.7\\
    & \checkmark & $[4,10]$ & 77.3\\
    & \checkmark & $[5,10]$ & 76.9\\
    \cline{2-4}
    & \checkmark & $[3,6]$ & 76.8\\
    & \checkmark & $[3,8]$ & 77.5\\
    & \checkmark & $[3,12]$ & 77.1\\
    & \checkmark & $[3,14]$ & 76.7\\
    \cline{2-4}
    & \checkmark & $[3,10]$ & \textbf{78.1}\\
    & $\times$ & $[3,10]$ & 76.3\\
    \hline
    \hline
    \end{tabular}
    }
    \smallskip
    \caption{Ablation study of range of incoherence lengths and jittering. The range of $l_{inc}$ is illustrated as $[l_{inc}^{min},l_{inc}^{max}]$}
    \label{Table:Range}
\end{table}

\subsection{Ablation studies}
In this section, we justify the design of our proposed VID by ablation studies. We first illustrate the optimal range of the incoherence length, and then conduct experiments utilizing different sub-tasks. Our proposed incoherence detection is additionally evaluated with various backbones compared with previous coherence-based methods. Except as otherwise specified, our ablation studies are conducted with R(2+1)D \cite{tran2018closer} pre-trained on UCF101.

\textbf{Range of the incoherence length.}
We first explore the best range of incoherence length $l_{inc}$. Illustrated as Table~\ref{Table:Range}, the experiments are conducted by changing either the lower bound $l_{inc}^{min}$ or the upper bound $l_{inc}^{max}$ of the range. As $l_{inc}^{min}$ increases from 2, we can also observe an improvement of performance from $76.7\%$ peaking at $78.1\%$ with $l_{inc}^{min}=3$, while the performance begins to decrease when $l_{inc}^{min}$ further increases. When $l_{inc}^{min}$ is smaller than 3, the incoherence between sub-clips are too difficult for the network to identify. The further increase of lower bound decreases the variety of incoherence length, leading to a drop in performance. Similar to the lower bound, when the upper bound of incoherence length increases from $l_{inc}^{max}=6$, the performance of VID rises consistently from $76.8\%$ which reaches a climax when $l_{inc}^{max}=10$, whereas a decreasing performance is observed as upper bound further increases, dropping from $78.1\%$ with $l_{inc}^{max}=10$ to $76.7\%$ with $l_{inc}^{max}=14$. As $l_{inc}^{max}$ increases, the sample range of incoherence becomes more abundant, while the incoherence becomes too obvious when $l_{inc}^{max}>10$. This observation indicates that an inappropriate range of $l_{inc}$ can result in too vague or too obvious incoherence, which leads to inferior performance. We thus set the range of $l_{inc}$ as $[3,10]$ in the following experiments.

\begin{table}[t]
    \centering
    \resizebox{.8\linewidth}{!}{
    \begin{tabular}{l|c c c|c}
    \hline
    \hline
    Sub-tasks & LoD / $\alpha$ & LeD / $\beta$ & ICL / $\lambda$ & UCF101(\%) \\
    \hline
    Random Init & - & - & - & 56.7\\
    \hline
    LoD & 1 & - & - & 75.4\\
    LeD &- & 1 & - & 70.9\\
    ICL &- & - & 1 & 72.1\\
    \hline
    LoD+LeD & 1 & 0.1 & - & 77.3\\
    LoD+ICL & 1 & - & 0.1 & 76.9\\
    LeD+ICL & - & 1 & 0.1 & 71.8\\
    \hline
    LoD+LeD+ICL& 1 & 0.1 & 0.1 & \textbf{78.1}\\
    \hline
    \hline
    \end{tabular}
    }
    \smallskip
    \caption{Ablation study of different sub-tasks.}
    \label{Table:Subtask}
\end{table}

\textbf{Different sub-tasks.}
We further evaluate the performance of different sub-tasks. As shown in Table~\ref{Table:Subtask}, when utilizing a single sub-task, we observe that networks pre-trained with any sub-task significantly exceed random initialization with a relative improvement of more than $25.0\%$. The network with LoD obtains the highest performance of $75.4\%$ on UCF101, which justifies the effectiveness of LoD. The network with ICL also achieves a competitive performance of $72.1\%$. However, when optimizing with only LeD, the network is required to directly predict the incoherence length without locating it. Therefore, the network can not fully leverage incoherence detection in videos, leading to inferior performance of $70.9\%$.

As for arbitrary pairs of sub-tasks, we observe that LoD-based pairs (LoD+LeD and LoD+ICL) surpass the single LoD with noticeable margins of more than $2.0\%$. This justifies the effectiveness of LeD and ICL as additional objectives. We also observe an inferior performance of the LeD-based pair (LeD+ICL), which aligns with the performance of the network with a single LeD. 

\begin{figure}[t]
    \centering{
    \includegraphics[width=.9\linewidth]{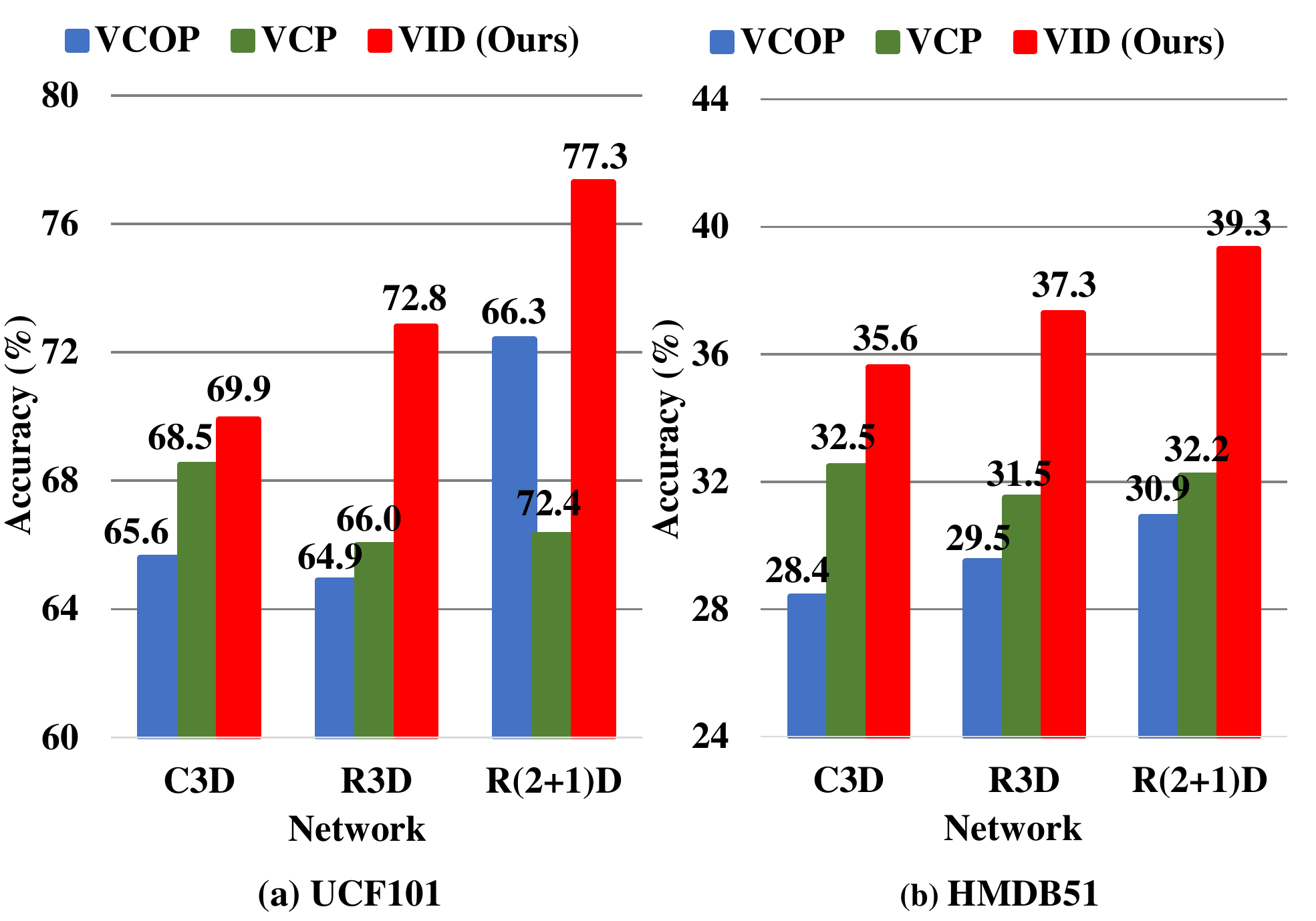}}
    \hfill
    \caption{Comparison with coherence-based methods.}
    \label{Figure:Backbone}
\end{figure}


\textbf{Comparison with coherence-based methods.}
To justify the effectiveness of incoherence detection, we evaluate our VID without additional ICL sub-task compared with previous coherence-based methods \cite{luo2020video,xu2019VCOP} utilizing order prediction. Illustrated in Figure~\ref{Figure:Backbone}, it is seen that our VID outperforms previous coherence-based methods across various backbones on different datasets. On UCF101, VID exceeds the previous VCOP \cite{xu2019VCOP} and VCP \cite{luo2020video} by $4.3\%$-$7.9\%$ and $1.4\%$-$11.0\%$, respectively. On HMDB51, the proposed VID surpasses VCOP \cite{xu2019VCOP} and VCP \cite{luo2020video} by over $10\%$ relatively. The improvement indicates that incoherence detection requires a more comprehensive understanding of videos compared with frame order reasoning.


\subsection{Evaluating self-supervised representation}

\begin{table}
    \centering
    \resizebox{\linewidth}{!}{
    \begin{tabular}{l c c c}
    \hline
    \hline
    Method & pre-train & UCF101(\%) & HMDB51(\%)\\
    \hline
    C3D (PRP \cite{Yao_2020PRP}) & UCF101 & 69.1 & 34.5\\
    C3D (PMAS \cite{wang2019PMAS}) & K-400 & 58.8 & 32.6\\
    C3D (RTT \cite{jenni2020RTT}) & K-600 & 69.9 & \textbf{39.6}\\
    C3D (Ours) & UCF101 & \textbf{70.2$\pm$0.5} & 37.7$\pm$0.7\\
    \hline
    R3D (ST-puzzle \cite{Kim2019STpuzzle}) & K-400 & 58.8 & 32.6\\
    R3D (PRP \cite{Yao_2020PRP}) & UCF101 & 66.5 & 29.7\\
    R3D (RTT \cite{jenni2020RTT}) & UCF101 & \textbf{77.3} & \textbf{47.5}\\
    R3D (Ours) & UCF101 & 73.6$\pm$0.5 & 38.0$\pm$0.6\\
    \hline
    R(2+1)D (PRP \cite{Yao_2020PRP}) & UCF101 & 72.1 & 35.0\\
    R(2+1)D (PP \cite{wang2020PP}) & K-400 & 77.1 & 36.6\\
    R(2+1)D (RTT \cite{jenni2020RTT}) & UCF101 & \textbf{81.6} & \textbf{46.4}\\
    R(2+1)D (Ours) & UCF101 & 78.1$\pm$0.6 & 40.1$\pm$0.6\\
    R(2+1)D (Ours) & K-400 & 78.5$\pm$0.4 & 41.5$\pm$0.5\\
    \hline
    \hline
    \end{tabular}
    }
    \smallskip
    \caption{Performance of action recognition compared with previous methods. RTT \cite{jenni2020RTT} is a SOTA method utilizing multiple data transformations. Results are the average of three evaluations (mean$\pm$std).}
    \label{Table:sotaAR}
\end{table}

\textbf{Action recognition.}
To verify the effectiveness of our proposed VID, we evaluate our VID with different backbones on action recognition which is a primary downstream task adopted in prior works \cite{wang2020PP,Yao_2020PRP,jenni2020RTT}. For action recognition, the network is initialized with the weights of the pre-trained model while the fully-connected layer is randomly initialized. The whole network is trained using the cross-entropy loss with an initial learning rate of 0.003. Other augmentation and parameter settings are the same as the pre-training stage. For testing, following the evaluation protocol of previous works \cite{wang2020PP,Yao_2020PRP}, we uniformly sample 10 clips from each video followed by a center crop. The final predictions for each video are the average result of all sampled clips. 

As shown in Table~\ref{Table:sotaAR}, our proposed VID achieves state-of-the-art (SOTA) results compared with self-supervised spatio-temporal reasoning methods which utilize a single data transformation. For C3D, our proposed VID outperforms the SOTA PRP method \cite{Yao_2020PRP} by $1.1\%$ on UCF101 and $3.2\%$ on HMDB51. For R3D, VID also exceeds PRP and ST-Puzzle \cite{Kim2019STpuzzle}, which are the previous SOTA method on UCF101 and HMDB51, with noticeable margins of $7.1\%$ on UCF101 and $5.4\%$ on HMDB51. With R(2+1)D pre-trained on UCF101, VID further surpasses the previous SOTA method PP \cite{wang2020PP} by $1.0\%$ on UCF101 and $3.1\%$ on HMDB51. When pre-trained on Kinetics-400, the margins of improvement further expand to $1.4\%$ and $4.5\%$, respectively. Provided by the illustrated results, VID can learn more abundant spatio-temporal representations compared with previous single-transformation methods.

In Table~\ref{Table:sotaAR}, we also include the results of RTT \cite{jenni2020RTT} which assembles multiple transformations, leading to superior performance compared with single-transformation methods. Nevertheless, for C3D, VID is the only single-transformation method that outperforms RTT by $0.3\%$ on UCF101 and provides competitive performance on HMDB51. It is possible to further improve the performance of ensemble-based methods by including our VID, while we mainly focus on leveraging video coherence by using a single temporal transformation in this work.

\begin{table}
    \centering
    \resizebox{\linewidth}{!}{
    \begin{tabular}{l c c c c c}
    \hline
    \hline
    Method & Top1 & Top5 & Top10 & Top20 & Top50 \\
    \hline
    C3D (PRP \cite{Yao_2020PRP}) & 23.2 & 38.1 & 46.0 & 55.7 & 68.4\\
    C3D (PP \cite{wang2020PP}) & 20.0 & 37.4 & 46.9 & 58.5 & 73.1\\
    C3D (Ours) & \textbf{26.9} & \textbf{43.6} & \textbf{53.6} & \textbf{63.8} & \textbf{78.2}\\
    \hline
    R3D (PRP \cite{Yao_2020PRP}) & 22.8 & 38.5 & 46.7 & 55.2 & 69.1\\
    R3D (PP \cite{wang2020PP}) & 19.9 & 36.2 & 46.1 & 55.6 & 69.8\\
    R3D (RTT \cite{jenni2020RTT}) & 26.1 & \textbf{48.5} & \textbf{59.1} & \textbf{69.6} & \textbf{82.8}\\
    R3D (Ours) & \textbf{26.4} & 44.5 & 54.1 & 63.9 & 78.2\\
    \hline
    R(2+1)D (PRP \cite{Yao_2020PRP}) & 20.3 & 34.0 & 41.9 & 51.7 & 64.2\\
    R(2+1)D (PP \cite{wang2020PP}) & 17.9 & 34.3 & 44.6 & 55.5 & 72.0\\
    R(2+1)D (Ours) & \textbf{22.0} & \textbf{40.4} & \textbf{51.2} & \textbf{61.8} & \textbf{74.7}\\
    \hline
    \hline
    \end{tabular}
    }
    \smallskip
    \caption{Performance of video retrieval on UCF101.}
    \label{Table:sotaVR-UCF}
\end{table}

\begin{table}
    \centering
    \resizebox{\linewidth}{!}{
    \begin{tabular}{l c c c c c}
    \hline
    \hline
    Method & Top1 & Top5 & Top10 & Top20 & Top50 \\
    \hline
    C3D (PRP \cite{Yao_2020PRP}) & 10.5 & 27.2 & 40.4 & 56.2 & 75.9\\
    C3D (PP \cite{wang2020PP}) & 8.0 & 25.2 & 37.8 & 54.4 & \textbf{77.5}\\
    C3D (Ours) & \textbf{11.6} & \textbf{29.6} & \textbf{43.3} & \textbf{58.4} & 77.3\\
    \hline
    R3D (PRP \cite{Yao_2020PRP}) & 8.2 & 25.8 & 38.5 & 53.3 & 75.9\\
    R3D (PP \cite{wang2020PP}) & 8.2 & 24.2 & 37.3 & 53.3 & 74.5\\
    R3D (Ours) & \textbf{11.2} & \textbf{32.2} & \textbf{45.4} & \textbf{59.8} & \textbf{79.2}\\
    \hline
    R(2+1)D (PRP \cite{Yao_2020PRP}) & 8.2 & 25.3 & 36.2 & 51.0 & 73.0\\
    R(2+1)D (PP \cite{wang2020PP}) & 10.1 & 24.6 & 37.6 & 54.4 & \textbf{77.1}\\
    R(2+1)D (Ours) & \textbf{10.4} & \textbf{27.9} & \textbf{42.7} & \textbf{58.1} & 76.7\\
    \hline
    \hline
    \end{tabular}
    }
    \smallskip
    \caption{Performance of video retrieval on HMDB51}
    \label{Table:sotaVR-HMDB}
\end{table}

\textbf{Video retrieval.}
We further evaluate our VID on another downstream task of nearest-neighbour video retrieval, which evaluates the quality of features extracted by the self-supervised pre-trained model. To make a fair comparison, our evaluation follows the protocol of previous state-of-the-art methods \cite{wang2020PP,Yao_2020PRP}. All models are pre-trained on UCF101. Given ten 16-frame clips sampled from each video, their features are extracted from the last pooling layer of the pre-trained backbone model. During inference, frames of each clip are first resized to 128 × 171 and then centrally cropped 112 × 112. Clips in the testing split are utilized to query the Top $k$ nearest samples based on their corresponding features. Here we consider $k$ equal to 1, 5, 10, 20, 50.

As shown in Table~\ref{Table:sotaVR-UCF} and Table~\ref{Table:sotaVR-HMDB}, VID outperforms state-of-the-art methods PP \cite{wang2020PP} and PRP \cite{Yao_2020PRP} on most evaluation metrics of UCF101 and HMDB51 across all backbones with significant marginals (e.g. $1.7\%$-$4.4\%$ for Top1 on UCF101). Specifically, VID surpasses all previous methods on HMDB51 across all evaluation metrics except for Top50, with improvement ranging from $0.3\%$ to $8.1\%$. The significant improvement further justifies that our VID extracts more effective spatio-temporal representation for downstream tasks.

\section{Conclusion}
In this paper, we propose a novel self-supervised method based on video incoherence detection for video representation learning. The incoherent clip is generated as the concatenation of sub-clips sampled from the same video with incoherence between each other. By detecting the location and length of incoherence, the network can extract effective spatio-temporal features. The intra-video contrastive learning is developed to maximize the mutual information between sub-clips from the same raw video. Extensive experiments show that VID achieves state-of-the-art performance with significant margins compared with previous methods. The proposed VID reveals a new perspective to leverage video coherence for video representation learning.

\section{Appendix}
\begin{figure}[t]
    \centering{
    \includegraphics[width=.95\linewidth]{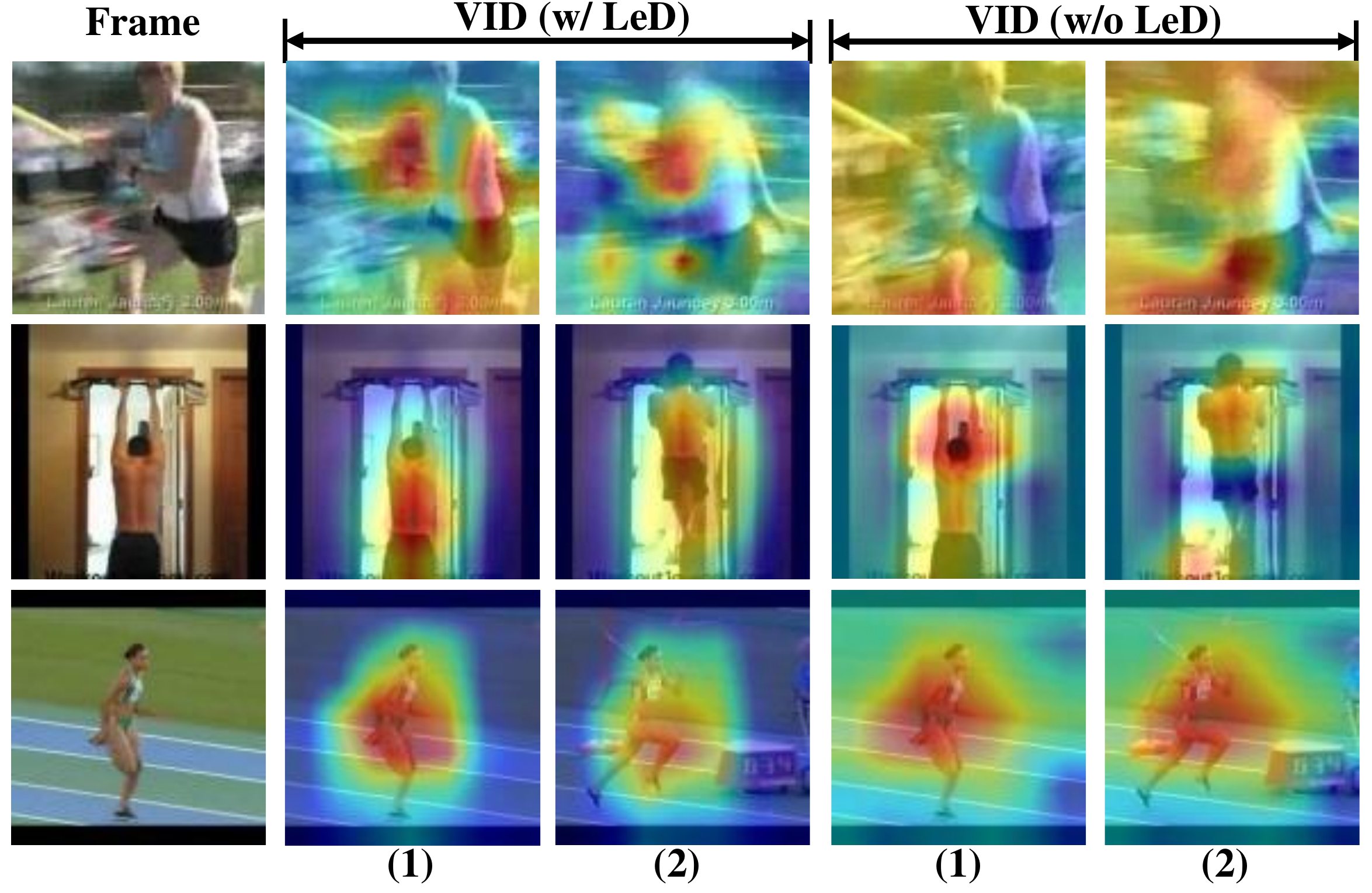}}
    \hfill
    \caption{Visualization of heat maps with/without LeD. The heat maps are generated from the last convolution layer based on Gradient-weighted Class Activation Mapping (Grad-CAM) \cite{GradCAM}. The indices below indicate the corresponding sub-clips which the frame belongs to.}
    \label{Figure:CAMVisual}
\end{figure}

\subsection{Implementation details}
In addition to experiment settings in our paper, we present the implementation details of our experiments. As mentioned in Sec.~4.1, color jittering is applied to each extracted incoherent clip along the temporal dimension. Specifically, we randomly modify the brightness, contrast, saturation and hue for each frame in the incoherent clip, whose augmentation ranges are $[0.2,1.8]$, $[0.2,1.8]$, $[0.2,1.8]$ and $[-0.2,0.2]$, respectively. Color jittering is enabled during the pre-training and fine-tuning stages and is disabled during evaluation. We conduct our experiments utilizing PyTorch \cite{paszke2017automatic} with two NVIDIA Tesla P100. \textit{Our code is available in the appendix.}

\subsection{Heat map visualization}
We visualize the heat map of extracted representation to justify the effectiveness of LeD task as shown in Figure~\ref{Figure:CAMVisual}. To validate the regularization effect of LeD, we additionally present corresponding heat maps from the network pre-trained without LeD. As shown in the last two rows, when there is subtle difference between scenes of sub-clips, networks
\begin{figure}[t!]
    \center{
    \includegraphics[width=.95\linewidth]{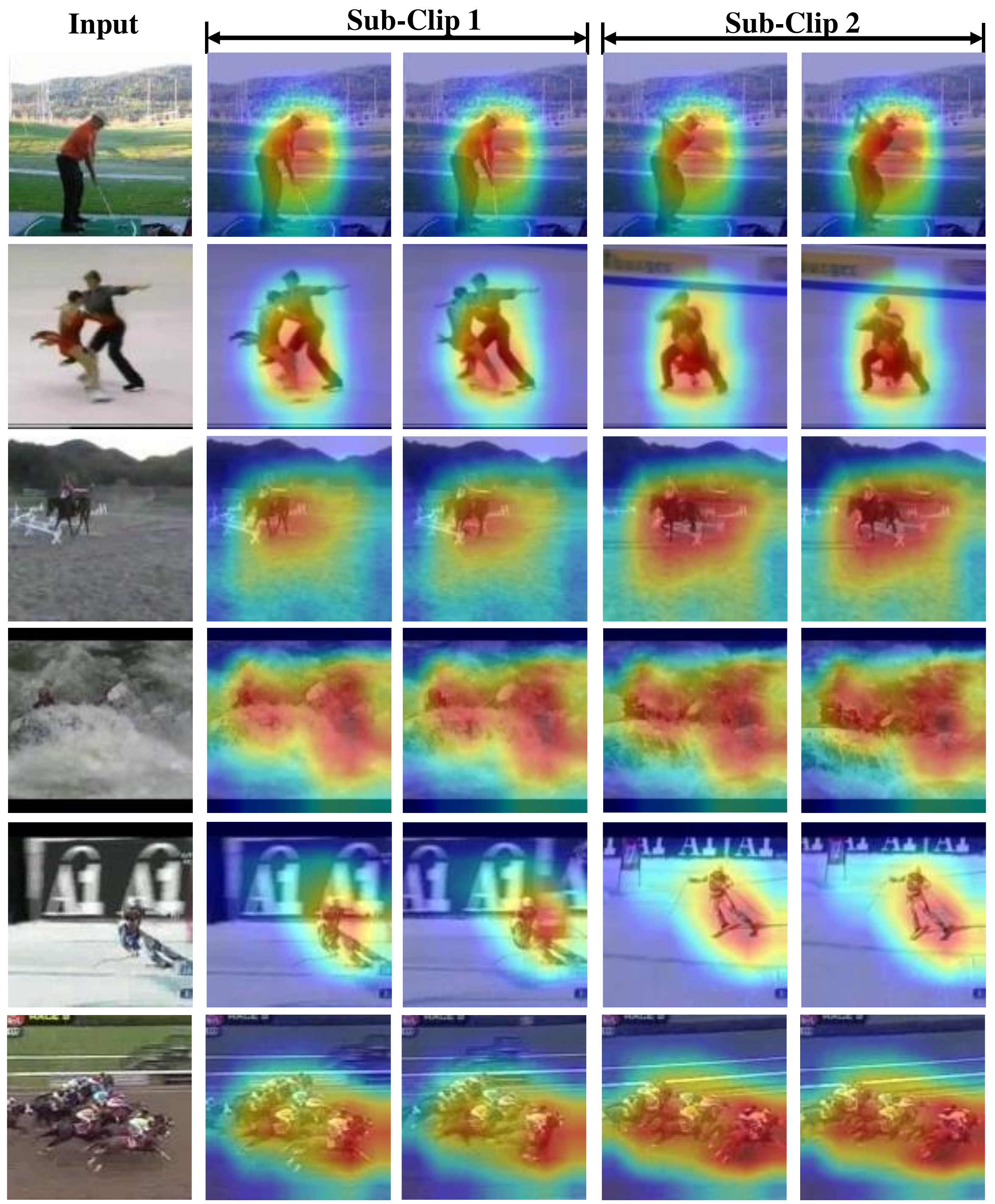}}
    \hfill
    \caption{Heat map visualization of our VID. The column ``Input'' contains original frames from the test samples. The following frames are sampled from different sub-clips.}
    \label{Figure:Sup-CAM}
\end{figure}
pre-trained with or without LeD both focus on the actors to detect abnormal motion caused by incoherence, which proves that incoherence detection requires motion understanding. When scenes change intensively as shown in the first row, the network pre-trained with LeD maintains its concentration on motion areas, while the network without LeD is distracted by the dynamic scene. This observation justifies that the utilization of LeD increases the robustness of VID towards the severe changes of low-level information, which avoids trivial learning. 

\begin{figure*}[!t]
    \centering{
    \includegraphics[width=1\textwidth]{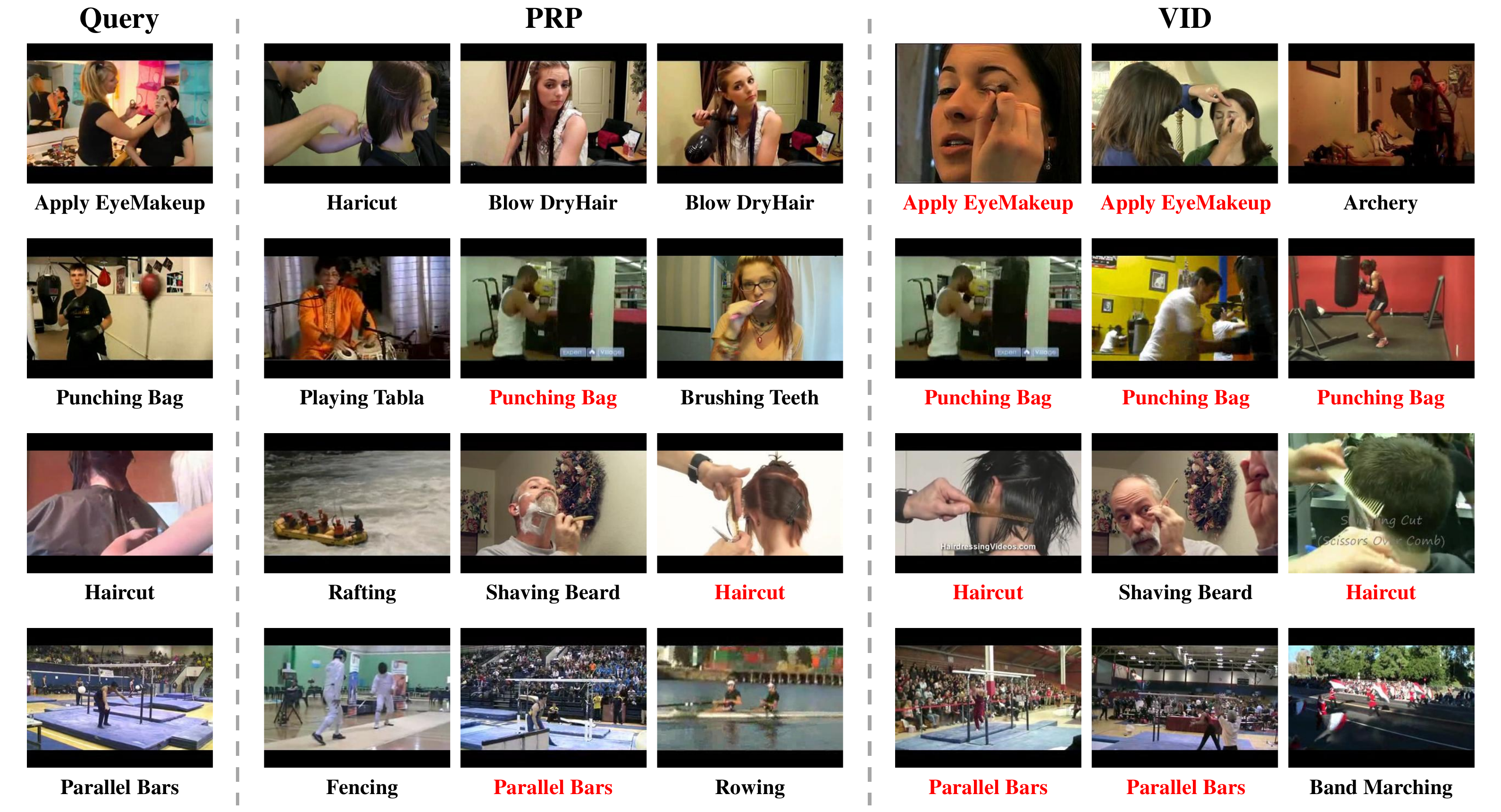}}
    \hfill
    \caption{Visualization of video retrieval results of our VID and previous PRP \cite{Yao_2020PRP}. The figures in the first column are queries. For each query, we present Top 3 retrieval results of our VID and the previous state-of-the-art PRP \cite{Yao_2020PRP}. Action classes in red represent the correct retrieval results.}
    \label{Figure:Sup-VR}
\end{figure*}

In addition to Figure~\ref{Figure:CAMVisual}, we present more heat map visualization of our VID to justify its effectiveness. As shown in Figure~\ref{Figure:Sup-CAM}, each row is augmented frames extracted from the test sample of UCF101 \cite{soomro2012ucf101}. The first column is the original frame representing the input sample. The following columns are heat maps visualization by Grad-CAM \cite{GradCAM} of different sub-clips. 

The heat maps provided in Figure~\ref{Figure:Sup-CAM} justify our assumption that incoherence detection requires an understanding of motion in videos. For example, given the golf swing in the first row, the network pre-trained with our VID concentrates on the upper body of the actor to detect the movements for incoherence detection. Additionally, as illustrated in the lower three rows in Figure~\ref{Figure:Sup-CAM}, when there are intensive changes between scenes of different sub-clip, our VID can maintain its concentration on the motion areas to detect incoherence. For example, given input indicating horse racing in the last row, the network pre-trained with our VID continuously focus on the horses and riders regardless of the intensive changes of backgrounds.

\subsection{Video retrieval results}
We present multiple examples of video retrieval results in comparison with the previous state-of-the-art method PRP \cite{Yao_2020PRP}. Following the evaluation protocols in previous works \cite{Yao_2020PRP,xu2019VCOP}, we extract the features from the last pooling layer of the pre-trained network. For PRP \cite{Yao_2020PRP}, we evaluate its performance with its provided pre-trained model. Both our VID and PRP \cite{Yao_2020PRP} are evaluated with ResNet18, which is the 18-layer variant of R3D \cite{tran2018closer}. Illustrated as Figure~\ref{Figure:Sup-VR}, our proposed VID provide more reasonable results compared to previous PRP \cite{Yao_2020PRP}. For example, given the query of applying eye makeup as shown in the first row, our VID retrieves two samples of the same action classes as the query among Top 3, while PRP retrieves samples that belong to similar-yet-incorrect actions, such as haircut and blow-dry hair. The retrieval results indicate that the network pre-trained with our VID obtains a more comprehensive understanding of videos compared to previous methods.

\clearpage
{\small
\bibstyle{aaai22}
\bibliography{aaai22}

\begin{thebibliography}{42}
\providecommand{\natexlab}[1]{#1}

\bibitem[{Alwassel et~al.(2019)Alwassel, Mahajan, Korbar, Torresani, Ghanem,
  and Tran}]{alwassel2019self}
Alwassel, H.; Mahajan, D.; Korbar, B.; Torresani, L.; Ghanem, B.; and Tran, D.
  2019.
\newblock Self-supervised learning by cross-modal audio-video clustering.
\newblock \emph{arXiv preprint arXiv:1911.12667}.

\bibitem[{Ando, Zhang, and Bartlett(2005)}]{ando2005framework}
Ando, R.~K.; Zhang, T.; and Bartlett, P. 2005.
\newblock A framework for learning predictive structures from multiple tasks
  and unlabeled data.
\newblock \emph{Journal of Machine Learning Research}, 6(11).

\bibitem[{Bachman, Hjelm, and Buchwalter(2019)}]{Phillip2019MM}
Bachman, P.; Hjelm, R.~D.; and Buchwalter, W. 2019.
\newblock Learning Representations by Maximizing Mutual Information Across
  Views.
\newblock In Wallach, H.; Larochelle, H.; Beygelzimer, A.; d\textquotesingle
  Alch\'{e}-Buc, F.; Fox, E.; and Garnett, R., eds., \emph{Advances in Neural
  Information Processing Systems}, volume~32. Curran Associates, Inc.

\bibitem[{Bottou(2010)}]{bottou2010SGD}
Bottou, L. 2010.
\newblock Large-scale machine learning with stochastic gradient descent.
\newblock In \emph{Proceedings of COMPSTAT'2010}, 177--186. Springer.

\bibitem[{Caruana and de(1997)}]{Caruana1997Promoting}
Caruana, R.; and de, V. 1997.
\newblock Promoting Poor Features to Supervisors: Some Inputs Work Better as
  Outputs.
\newblock In Mozer, M.~C.; Jordan, M.; and Petsche, T., eds., \emph{Advances in
  Neural Information Processing Systems}, volume~9. MIT Press.

\bibitem[{Devlin et~al.(2018)Devlin, Chang, Lee, and
  Toutanova}]{devlin2018bert}
Devlin, J.; Chang, M.-W.; Lee, K.; and Toutanova, K. 2018.
\newblock Bert: Pre-training of deep bidirectional transformers for language
  understanding.
\newblock \emph{arXiv preprint arXiv:1810.04805}.

\bibitem[{{Dwibedi} et~al.(2019){Dwibedi}, {Aytar}, {Tompson}, {Sermanet}, and
  {Zisserman}}]{Dwibedi2019TCC}
{Dwibedi}, D.; {Aytar}, Y.; {Tompson}, J.; {Sermanet}, P.; and {Zisserman}, A.
  2019.
\newblock Temporal Cycle-Consistency Learning.
\newblock In \emph{2019 IEEE/CVF Conference on Computer Vision and Pattern
  Recognition (CVPR)}, 1801--1810.

\bibitem[{Fernando et~al.(2017)Fernando, Bilen, Gavves, and
  Gould}]{fernando2017O3D}
Fernando, B.; Bilen, H.; Gavves, E.; and Gould, S. 2017.
\newblock Self-supervised video representation learning with odd-one-out
  networks.
\newblock In \emph{Proceedings of the IEEE conference on computer vision and
  pattern recognition}, 3636--3645.

\bibitem[{{Gan} et~al.(2018){Gan}, {Gong}, {Liu}, {Su}, and
  {Guibas}}]{Gan2018Geometry}
{Gan}, C.; {Gong}, B.; {Liu}, K.; {Su}, H.; and {Guibas}, L.~J. 2018.
\newblock Geometry Guided Convolutional Neural Networks for Self-Supervised
  Video Representation Learning.
\newblock In \emph{2018 IEEE/CVF Conference on Computer Vision and Pattern
  Recognition}, 5589--5597.

\bibitem[{Han, Xie, and Zisserman(2019)}]{han2019DPC}
Han, T.; Xie, W.; and Zisserman, A. 2019.
\newblock Video representation learning by dense predictive coding.
\newblock In \emph{Proceedings of the IEEE/CVF International Conference on
  Computer Vision Workshops}, 0--0.

\bibitem[{Han, Xie, and Zisserman(2020)}]{Han2020MemDPC}
Han, T.; Xie, W.; and Zisserman, A. 2020.
\newblock Memory-Augmented Dense Predictive Coding for Video Representation
  Learning.
\newblock In Vedaldi, A.; Bischof, H.; Brox, T.; and Frahm, J.-M., eds.,
  \emph{Computer Vision -- ECCV 2020}, 312--329. Cham: Springer International
  Publishing.
\newblock ISBN 978-3-030-58580-8.

\bibitem[{Hara, Kataoka, and Satoh(2018)}]{hara2018can}
Hara, K.; Kataoka, H.; and Satoh, Y. 2018.
\newblock Can spatiotemporal 3d cnns retrace the history of 2d cnns and
  imagenet?
\newblock In \emph{Proceedings of the IEEE conference on Computer Vision and
  Pattern Recognition}, 6546--6555.

\bibitem[{He et~al.(2020)He, Fan, Wu, Xie, and Girshick}]{he2020momentum}
He, K.; Fan, H.; Wu, Y.; Xie, S.; and Girshick, R. 2020.
\newblock Momentum contrast for unsupervised visual representation learning.
\newblock In \emph{Proceedings of the IEEE/CVF Conference on Computer Vision
  and Pattern Recognition}, 9729--9738.

\bibitem[{Hjelm et~al.(2019)Hjelm, Fedorov, Lavoie{-}Marchildon, Grewal,
  Bachman, Trischler, and Bengio}]{Hjelm2019MIE}
Hjelm, R.~D.; Fedorov, A.; Lavoie{-}Marchildon, S.; Grewal, K.; Bachman, P.;
  Trischler, A.; and Bengio, Y. 2019.
\newblock Learning deep representations by mutual information estimation and
  maximization.
\newblock In \emph{7th International Conference on Learning Representations,
  {ICLR} 2019, New Orleans, LA, USA, May 6-9, 2019}. OpenReview.net.

\bibitem[{Jenni, Meishvili, and Favaro(2020)}]{jenni2020RTT}
Jenni, S.; Meishvili, G.; and Favaro, P. 2020.
\newblock Video representation learning by recognizing temporal
  transformations.
\newblock In \emph{Computer Vision--ECCV 2020: 16th European Conference,
  Glasgow, UK, August 23--28, 2020, Proceedings, Part XXVIII 16}, 425--442.
  Springer.

\bibitem[{Kay et~al.(2017)Kay, Carreira, Simonyan, Zhang, Hillier,
  Vijayanarasimhan, Viola, Green, Back, Natsev et~al.}]{kay2017kinetics}
Kay, W.; Carreira, J.; Simonyan, K.; Zhang, B.; Hillier, C.; Vijayanarasimhan,
  S.; Viola, F.; Green, T.; Back, T.; Natsev, P.; et~al. 2017.
\newblock The kinetics human action video dataset.
\newblock \emph{arXiv preprint arXiv:1705.06950}.

\bibitem[{Kim, Cho, and Kweon(2019)}]{Kim2019STpuzzle}
Kim, D.; Cho, D.; and Kweon, I.~S. 2019.
\newblock Self-Supervised Video Representation Learning with Space-Time Cubic
  Puzzles.
\newblock \emph{Proceedings of the AAAI Conference on Artificial Intelligence},
  33(01): 8545--8552.

\bibitem[{Kuehne et~al.(2011)Kuehne, Jhuang, Garrote, Poggio, and
  Serre}]{kuehne2011hmdb}
Kuehne, H.; Jhuang, H.; Garrote, E.; Poggio, T.; and Serre, T. 2011.
\newblock HMDB51: A Large Video Database for Human Motion Recognition.
\newblock In \emph{Proceedings of the IEEE International Conference on Computer
  Vision}, 2556--2563.

\bibitem[{Lan et~al.(2019)Lan, Chen, Goodman, Gimpel, Sharma, and
  Soricut}]{lan2019albert}
Lan, Z.; Chen, M.; Goodman, S.; Gimpel, K.; Sharma, P.; and Soricut, R. 2019.
\newblock Albert: A lite bert for self-supervised learning of language
  representations.
\newblock \emph{arXiv preprint arXiv:1909.11942}.

\bibitem[{Lee et~al.(2017)Lee, Huang, Singh, and Yang}]{lee2017sortsequence}
Lee, H.-Y.; Huang, J.-B.; Singh, M.; and Yang, M.-H. 2017.
\newblock Unsupervised representation learning by sorting sequences.
\newblock In \emph{Proceedings of the IEEE International Conference on Computer
  Vision}, 667--676.

\bibitem[{Lorre et~al.(2020)Lorre, Rabarisoa, Orcesi, Ainouz, and
  Canu}]{lorre2020temporal}
Lorre, G.; Rabarisoa, J.; Orcesi, A.; Ainouz, S.; and Canu, S. 2020.
\newblock Temporal contrastive pretraining for video action recognition.
\newblock In \emph{Proceedings of the IEEE/CVF Winter Conference on
  Applications of Computer Vision}, 662--670.

\bibitem[{Luo et~al.(2020)Luo, Liu, Zhou, Yang, Ma, Ye, and
  Wang}]{luo2020video}
Luo, D.; Liu, C.; Zhou, Y.; Yang, D.; Ma, C.; Ye, Q.; and Wang, W. 2020.
\newblock Video Cloze Procedure for Self-Supervised Spatio-Temporal Learning.
\newblock In \emph{The Thirty-Fourth {AAAI} Conference on Artificial
  Intelligence, {AAAI} 2020, The Thirty-Second Innovative Applications of
  Artificial Intelligence Conference, {IAAI} 2020, The Tenth {AAAI} Symposium
  on Educational Advances in Artificial Intelligence, {EAAI} 2020, New York,
  NY, USA, February 7-12, 2020}, 11701--11708. {AAAI} Press.

\bibitem[{Luo et~al.(2017)Luo, Peng, Huang, Alahi, and
  Fei-Fei}]{luo2017unsupervised}
Luo, Z.; Peng, B.; Huang, D.-A.; Alahi, A.; and Fei-Fei, L. 2017.
\newblock Unsupervised learning of long-term motion dynamics for videos.
\newblock In \emph{Proceedings of the IEEE conference on computer vision and
  pattern recognition}, 2203--2212.

\bibitem[{Misra and Maaten(2020)}]{Misra_2020_PIR}
Misra, I.; and Maaten, L. v.~d. 2020.
\newblock Self-Supervised Learning of Pretext-Invariant Representations.
\newblock In \emph{Proceedings of the IEEE/CVF Conference on Computer Vision
  and Pattern Recognition (CVPR)}.

\bibitem[{Misra, Zitnick, and Hebert(2016)}]{misra2016shuffle}
Misra, I.; Zitnick, C.~L.; and Hebert, M. 2016.
\newblock Shuffle and learn: unsupervised learning using temporal order
  verification.
\newblock In \emph{European Conference on Computer Vision}, 527--544. Springer.

\bibitem[{Oord, Li, and Vinyals(2018)}]{oord2018representation}
Oord, A. v.~d.; Li, Y.; and Vinyals, O. 2018.
\newblock Representation learning with contrastive predictive coding.
\newblock \emph{arXiv preprint arXiv:1807.03748}.

\bibitem[{Paszke et~al.(2017)Paszke, Gross, Chintala, Chanan, Yang, DeVito,
  Lin, Desmaison, Antiga, and Lerer}]{paszke2017automatic}
Paszke, A.; Gross, S.; Chintala, S.; Chanan, G.; Yang, E.; DeVito, Z.; Lin, Z.;
  Desmaison, A.; Antiga, L.; and Lerer, A. 2017.
\newblock Automatic differentiation in PyTorch.
\newblock In \emph{NIPS-W}.

\bibitem[{{Selvaraju} et~al.(2017){Selvaraju}, {Cogswell}, {Das}, {Vedantam},
  {Parikh}, and {Batra}}]{GradCAM}
{Selvaraju}, R.~R.; {Cogswell}, M.; {Das}, A.; {Vedantam}, R.; {Parikh}, D.;
  and {Batra}, D. 2017.
\newblock Grad-CAM: Visual Explanations from Deep Networks via Gradient-Based
  Localization.
\newblock In \emph{2017 IEEE International Conference on Computer Vision
  (ICCV)}, 618--626.

\bibitem[{Soomro, Zamir, and Shah(2012)}]{soomro2012ucf101}
Soomro, K.; Zamir, A.~R.; and Shah, M. 2012.
\newblock UCF101: A dataset of 101 human actions classes from videos in the
  wild.
\newblock \emph{arXiv preprint arXiv:1212.0402}.

\bibitem[{Srivastava, Mansimov, and
  Salakhutdinov(2015)}]{Nitish2015frameprediction}
Srivastava, N.; Mansimov, E.; and Salakhutdinov, R. 2015.
\newblock Unsupervised Learning of Video Representations Using LSTMs.
\newblock In \emph{Proceedings of the 32nd International Conference on
  International Conference on Machine Learning - Volume 37}, ICML'15,
  843–852. JMLR.org.

\bibitem[{Sun et~al.(2019{\natexlab{a}})Sun, Baradel, Murphy, and
  Schmid}]{sun2019learning}
Sun, C.; Baradel, F.; Murphy, K.; and Schmid, C. 2019{\natexlab{a}}.
\newblock Learning video representations using contrastive bidirectional
  transformer.
\newblock \emph{arXiv preprint arXiv:1906.05743}.

\bibitem[{Sun et~al.(2019{\natexlab{b}})Sun, Myers, Vondrick, Murphy, and
  Schmid}]{sun2019videobert}
Sun, C.; Myers, A.; Vondrick, C.; Murphy, K.; and Schmid, C.
  2019{\natexlab{b}}.
\newblock Videobert: A joint model for video and language representation
  learning.
\newblock In \emph{Proceedings of the IEEE/CVF International Conference on
  Computer Vision}, 7464--7473.

\bibitem[{Tian et~al.(2020)Tian, Che, Bao, Zhai, and Gao}]{Yuan2020CGMUL}
Tian, Y.; Che, Z.; Bao, W.; Zhai, G.; and Gao, Z. 2020.
\newblock Self-supervised Motion Representation via Scattering Local Motion
  Cues.
\newblock In Vedaldi, A.; Bischof, H.; Brox, T.; and Frahm, J.-M., eds.,
  \emph{Computer Vision -- ECCV 2020}, 71--89. Cham: Springer International
  Publishing.
\newblock ISBN 978-3-030-58568-6.

\bibitem[{{Tran} et~al.(2015){Tran}, {Bourdev}, {Fergus}, {Torresani}, and
  {Paluri}}]{Tran2015C3D}
{Tran}, D.; {Bourdev}, L.; {Fergus}, R.; {Torresani}, L.; and {Paluri}, M.
  2015.
\newblock Learning Spatiotemporal Features with 3D Convolutional Networks.
\newblock In \emph{2015 IEEE International Conference on Computer Vision
  (ICCV)}, 4489--4497.

\bibitem[{Tran et~al.(2018)Tran, Wang, Torresani, Ray, LeCun, and
  Paluri}]{tran2018closer}
Tran, D.; Wang, H.; Torresani, L.; Ray, J.; LeCun, Y.; and Paluri, M. 2018.
\newblock A closer look at spatiotemporal convolutions for action recognition.
\newblock In \emph{Proceedings of the IEEE conference on Computer Vision and
  Pattern Recognition}, 6450--6459.

\bibitem[{Vondrick, Pirsiavash, and Torralba(2016)}]{Carl2016Generating}
Vondrick, C.; Pirsiavash, H.; and Torralba, A. 2016.
\newblock Generating Videos with Scene Dynamics.
\newblock In \emph{Proceedings of the 30th International Conference on Neural
  Information Processing Systems}, NIPS'16, 613–621. Red Hook, NY, USA:
  Curran Associates Inc.
\newblock ISBN 9781510838819.

\bibitem[{Wang et~al.(2019)Wang, Jiao, Bao, He, Liu, and Liu}]{wang2019PMAS}
Wang, J.; Jiao, J.; Bao, L.; He, S.; Liu, Y.; and Liu, W. 2019.
\newblock Self-supervised spatio-temporal representation learning for videos by
  predicting motion and appearance statistics.
\newblock In \emph{Proceedings of the IEEE/CVF Conference on Computer Vision
  and Pattern Recognition}, 4006--4015.

\bibitem[{Wang, Jiao, and Liu(2020)}]{wang2020PP}
Wang, J.; Jiao, J.; and Liu, Y.-H. 2020.
\newblock Self-supervised video representation learning by pace prediction.
\newblock In \emph{European Conference on Computer Vision}, 504--521. Springer.

\bibitem[{Wu et~al.(2018)Wu, Xiong, Yu, and Lin}]{wu2018unsupervised}
Wu, Z.; Xiong, Y.; Yu, S.~X.; and Lin, D. 2018.
\newblock Unsupervised feature learning via non-parametric instance
  discrimination.
\newblock In \emph{Proceedings of the IEEE Conference on Computer Vision and
  Pattern Recognition}, 3733--3742.

\bibitem[{Xu et~al.(2019)Xu, Xiao, Zhao, Shao, Xie, and Zhuang}]{xu2019VCOP}
Xu, D.; Xiao, J.; Zhao, Z.; Shao, J.; Xie, D.; and Zhuang, Y. 2019.
\newblock Self-supervised spatiotemporal learning via video clip order
  prediction.
\newblock In \emph{Proceedings of the IEEE/CVF Conference on Computer Vision
  and Pattern Recognition}, 10334--10343.

\bibitem[{Yao et~al.(2021)Yao, Zhang, Qiu, Pan, and Mei}]{yao2021seco}
Yao, T.; Zhang, Y.; Qiu, Z.; Pan, Y.; and Mei, T. 2021.
\newblock SeCo: Exploring Sequence Supervision for Unsupervised Representation
  Learning.
\newblock In \emph{35th AAAI Conference on Artificial Intelligence}.

\bibitem[{Yao et~al.(2020)Yao, Liu, Luo, Zhou, and Ye}]{Yao_2020PRP}
Yao, Y.; Liu, C.; Luo, D.; Zhou, Y.; and Ye, Q. 2020.
\newblock Video Playback Rate Perception for Self-Supervised Spatio-Temporal
  Representation Learning.
\newblock In \emph{Proceedings of the IEEE/CVF Conference on Computer Vision
  and Pattern Recognition (CVPR)}.

\end{thebibliography}
}

\end{document}